\documentclass[conference]{IEEEtran}
\usepackage{times}
\usepackage{todonotes}
\usepackage{multicol}
\usepackage{graphics,graphicx,caption,float,subcaption,booktabs,xcolor,multirow,array,color,ifthen,tabu,colortbl,dblfloatfix,url,xparse,mathtools,algorithm,algorithmic,amssymb,xspace,nicefrac,microtype,amsmath,amsfonts,bm,ragged2e,tikz,stackengine,etoolbox,xpatch,enumerate,xstring,setspace,tabularx,makecell,changepage,cuted,wrapfig, sidecap,comment, bbding, placeins}
\usepackage{gensymb,comment}
\usepackage{cite} 

\usepackage[pagebackref=true,breaklinks=true,colorlinks=true,bookmarks=false,citecolor=blue]{hyperref}
\hypersetup{
colorlinks=true,
linkcolor=blue,
filecolor=magenta,      
citecolor=blue
}

\usepackage{siunitx}
\IEEEoverridecommandlockouts                              

\title{\LARGE \bf ClearDepth: Efficient Stereo Perception of Transparent Objects \\ for Robotic Manipulation}

\ifdefined\isanonymous
\else
\fi

\ifdefined\isanonymous
    \author{%
        Anonymous Authors\\
    \thanks{Affiliations withheld for double-blind review.
    }\\
    \thanks{
    }\\
    \thanks{
    }
    }
\else
    \author{
    Kaixin Bai$^{1,2}$, Huajian Zeng$^{2,3,4}$, Lei Zhang$^{1,2\dag}$, Yiwen Liu$^{2,3}$, Hongli Xu$^{3}$, Zhaopeng Chen$^{2}$, Jianwei Zhang$^{1}$ 
\thanks{\dag Corresponding author. zhanglei.cn.de@gmail.com, lei.zhang-1@studium.uni-hamburg.de}
    \thanks{$^{1}$TAMS (Technical Aspects of Multimodal Systems), Department of Informatics, University of Hamburg, Hamburg, Germany.}
\thanks{$^{2}$Agile Robots SE, Munich, Germany.}
\thanks{$^{3}$Technical University of Munich, Germany. 
}
\thanks{$^{4}$Mohamed Bin Zayed University of Artificial Intelligence (MBUZAI),
Abu Dhabi, UAE.
}
}
\fi

\begin{document}

\def\@onedot{\ifx\@let@token.\else.\null\fi\xspace}
\DeclareRobustCommand\onedot{\futurelet\@let@token\@onedot}
\newcommand{\figref}[1]{Fig\onedot~\ref{#1}}
\def\etal{\emph{et al}\onedot}
\newcommand{\secref}[1]{Sec\onedot~\ref{#1}}
\newcommand{\tabref}[1]{Tab\onedot~\ref{#1}}
\newcommand\ananye[1]{\textcolor{red}{#1}}
\makeatletter
\let\@oldmaketitle\@maketitle
\renewcommand{\@maketitle}{
\@oldmaketitle
    \centering
    \includegraphics[width=\linewidth]{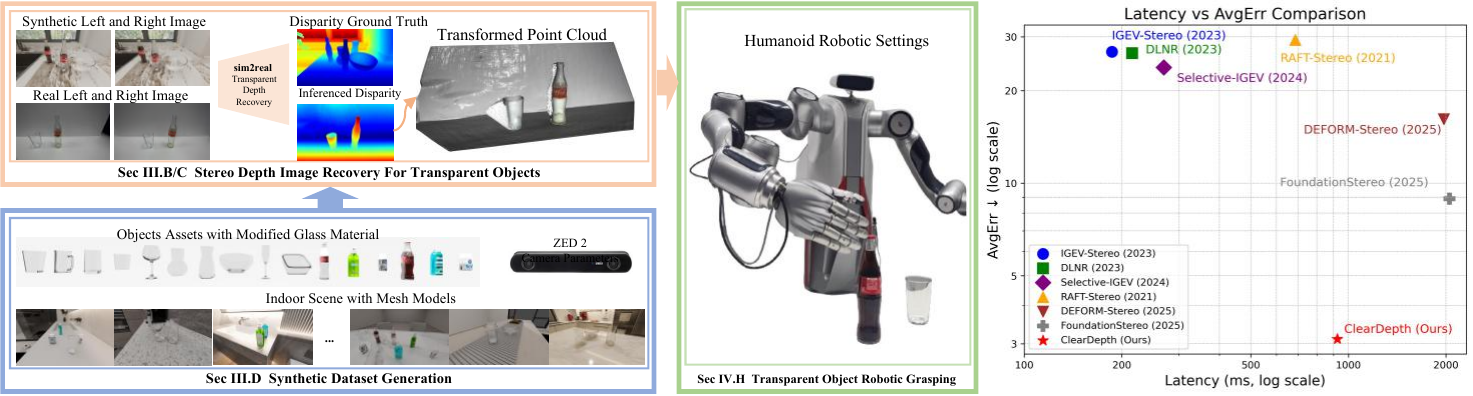}
    \captionof{figure}{%
    ClearDepth leverages structure-aware stereo matching and synthetic training data to bridge the Sim2Real gap in transparent object grasping, achieving superior speed–accuracy trade-offs.
    }
    \label{fig.cleardepth_overview}
\vspace{-2em}
\bigskip
}

\maketitle

\setcounter{figure}{1}

\begin{abstract}
Transparent object depth perception remains a major challenge in robotics and logistics due to the limitations of standard 3D sensors in capturing accurate depth on transparent and reflective surfaces. This affects applications relying on depth maps and point clouds, particularly in robotic manipulation. To address this, we propose ClearDepth, a vision transformer-based algorithm for stereo depth recovery of transparent objects, enhanced by a novel feature post-fusion module that refines depth estimation using structural visual features. To mitigate the high costs of stereo dataset collection, we introduce a physically realistic, domain-adaptive Sim2Real framework for efficient data generation. 
Our method outperforms state-of-the-art stereo matching approaches on transparent depth recovery. 
Furthermore, in transparent object grasping experiments, ClearDepth improves transparent-scene perception and achieves at least an 18\% higher grasp success rate compared to the state-of-the-art methods for transparent object manipulation. 
Our method demonstrates strong Sim2Real generalization, enabling precise depth perception of transparent objects for robotic applications in the real world. 
Dataset and project details are available at 
\ifdefined\isanonymous
\href{https://sites.google.com/view/cleardepth-anonymous/}{https://sites.google.com/view/cleardepth-anonymous/}.
\else
\href{https://sites.google.com/view/cleardepth/}{https://sites.google.com/view/cleardepth/}.
\fi
\end{abstract}

\section{INTRODUCTION}
\label{sec:intro}

Transparent objects, such as glass bottles and cups, are prevalent in domestic service robotics and logistics sorting scenarios. However, their inherent transparency, particularly the complex effects of refraction and reflection, poses significant challenges for visual perception and recognition~\cite{jiang2023robotic}. These perception limitations, in turn, constrain the robot’s ability to manipulate such objects effectively in real-world tasks.

Deep learning has played a critical role in understanding and modeling the complex geometrical features of transparent objects. To address these challenges, prior research has primarily focused on enhancing perception capabilities through deep learning, such as reconstructing depth from incomplete depth maps~\cite{li2023fdct,chen2023tode}, stereo visual perception~\cite{chen2023stereopose}, and multi-view approaches~\cite{wang2023mvtrans}. 
Despite notable progress, real-world applications still face difficulties in extracting reliable feature points due to inconsistencies in depth data input and the increased complexity of multi-view imaging systems. 
Transparent objects refract background textures, making structural features more critical than texture features for imaging and perception. 
To obtain stable feature points, some studies have explored extracting structural details~\cite{cao2021fakemix} or improving the precision of depth sensing hardware~\cite{shao2022polarimetric}. However, these approaches remain limited in effectiveness and generalization ability. 

Studies have shown that CNNs excel at texture recognition, while vision Transformers (ViTs) demonstrate superior capabilities in modeling shape features~\cite{tuli2021convolutional}. However, traditional ViTs typically downsample the input and rely on learnable upsampling to restore spatial resolution. While effective, this approach is computationally expensive and often lacks the ability to capture fine-grained details. RAFT-Stereo~\cite{lipson2021raft}, an extension of RAFT~\cite{teed2020raft}, applies optical flow techniques to stereo matching, improving generalization and robustness with its lightweight Gated Recurrent Unit (GRU) Network module, but struggles with global context extraction due to its CNN architecture. To address these limitations, models such as SegFormer~\cite{xie2021segformer} and DinoV2~\cite{oquab2023dinov2} enhance ViTs through cascaded architectures and multi-scale feature fusion, improving performance in depth estimation and semantic segmentation tasks. 
Transparent objects pose additional challenges due to their optical properties, which often cause background textures to become distorted. As a result, texture-based features become unreliable, and structural features become crucial for accurate perception. To better capture these structural cues, we design an efficient \textbf{cascaded ViT backbone} to extract contextual structural information, making it well-suited for modeling transparent object scenes. 
Moreover, conventional stereo matching networks typically rely on dot-product similarity for feature correspondence, which is ineffective in transparent object scenarios due to background refraction. 
To overcome this, we introduce a lightweight \textbf{post-fusion module} that incorporates structural feature priors into the Gated Recurrent Unit (GRU) update loop. This design improves structural awareness without introducing the computational overhead of cross-attention mechanisms. The whole pipeline is shown in Fig.~\ref{fig.cleardepth_overview}.

Accurate datasets are vital for deep learning on transparent objects, yet existing collection methods, such as pose markers~\cite{fang2022transcg,xu2021seeing}, opaque substitutes~\cite{sajjan2020clear}, and manual 3D modeling~\cite{chen2022clearpose}, are labor-intensive and yield noisy depth maps. To address this, simulation engines are increasingly used~\cite{chen2022clearpose,wang2023mvtrans}, though balancing realism and efficiency remains a challenge. We propose \textbf{SynClearDepth}, a synthetic dataset generated via a realistic data generation pipeline that supports direct model deployment on real-world sensors, providing instance segmentation, object poses, and depth maps.

In summary, our main contributions are:
\begin{enumerate}

\item An efficient stereo depth recovery network \textbf{ClearDepth} for transparent objects, featuring a cascaded ViT encoder for multi-scale structural feature extraction and a lightweight post-fusion module that integrates structural priors with appearance cues to achieve robust and efficient depth estimation.

\item {The demonstrated advancements over SOTA methods, as evidenced in stereo perception benchmarks and real-world scenarios, exhibit significant qualitative and quantitative enhancements in the robotic grasping of transparent objects in single-object and cluttered environments, underscoring our solution's superior effectiveness.}

\item SynClearDepth, a photo-realistic dataset for transparent object perception in grasping scenes, containing 14,091 stereo RGB images with ground-truth depth and segmentation labels. It aligns simulated with real sensor parameters and leverages domain randomization and adaptation to ensure diversity and robustness across different scenes and camera settings.
\end{enumerate}

\section{Related Work}\label{sec:relatedwork}

\subsection{Transparent Object Perception}

Robotic perception of transparent objects remains challenging due to their low contrast and complex light interactions, which affect sensor accuracy in determining position and shape. Traditional RGB and RGB-D cameras struggle with these objects, as they rely on intensity data and overlook optical properties.
To address this, research has explored polarized cameras, which reduce reflections and enhance contrast~\cite{shao2022polarimetric,kalra2020deep}. However, their high cost limits widespread adoption. Alternative approaches include CNN-transformer-based models for tracking~\cite{garigapati2023transparent}, Sim2Real techniques leveraging synthetic datasets~\cite{lukezic2022trans2k}, and alpha-matting methods for transparent object segmentation~\cite{cai2023transmatting}.
For robotic manipulation and pose estimation, multi-task perception models have been introduced~\cite{fang2022transcg,chen2022clearpose,dai2022domain}. Depth recovery remains particularly challenging due to light refraction and reflection. Methods such as NeRF and volumetric rendering aid surface reconstruction~\cite{li2023neto,dai2023graspnerf,li2020through}, while stereo and multi-view techniques improve depth estimation~\cite{chen2023stereopose,zhang2023transnet,xu2021seeing}.
These approaches leverage various sensors, including RGB-D, stereo vision, and multi-view systems, to enhance transparent object perception~\cite{chen2023tode,li2023fdct,fang2022transcg,chen2022clearpose,dai2022domain,xu2021seeing,chen2023stereopose,wang2023mvtrans,dai2023graspnerf,shao2022polarimetric}. Advances in deep learning and sensor technologies continue to drive improvements in accuracy and reliability.

\subsection{Deep Learning-based Stereo Depth Recovery}
Deep learning-based stereo matching methods have recently outperformed traditional approaches, with 2D convolutional models~\cite{mayer2016large,xu2020aanet} offering simplicity and efficiency. These models achieve high accuracy even on limited computational resources, making them suitable for engineering applications, though they still require improvements in accuracy and robustness due to 3D cost space constraints.
3D convolutional networks~\cite{cao2019gcnet,chang2018pyramid} provide better interpretability and higher disparity map accuracy but require optimization due to their computational demands. 
STTR~\cite{li2021revisiting}, inspired by SuperGlue~\cite{sarlin2020superglue}, uses transformers with positional embedding and attention mechanisms for binocular dense matching, producing disparity and depth maps. However, these methods are computationally intensive and slow in inference, limiting their suitability for high-resolution images and downstream robotic tasks.

\subsection{Transparent Object Datasets}

Recent works~\cite{depth_anything_v2, bochkovskii2024depth} show that real-world datasets degrade model performance due to label noise, while synthetic data with precise labels, enhance model performance. 
However, synthetic datasets across different data domains remain scarce in the open-source community. 
Ray-tracing renderers have narrowed the sim2real gap, making domain differences the main bottleneck in model generalization.
Existing synthetic datasets, such as ~\cite{sajjan2020clear,zhu2021rgb,xu2021seeing,ichnowski2021dex,shi2024asgrasp,dai2022domain}, typically feature transparent objects on desktops. 
However, these datasets lack the complexity necessary for generalization to real-world scenarios like kitchens, bedrooms, and offices, where service and humanoid robots operate. 
Moreover, these datasets often require extensive pre- or post-processing, such as segmentation or background reconstruction, which is impractical for end-to-end algorithms crucial to embodied intelligence applications. 
Datasets using HDRI backgrounds~\cite{wang2023mvtrans} lack depth labels, which hinders generalization, especially in zero-shot tasks. Others simulate Realsense cameras~\cite{dai2022domain,shi2024asgrasp}, but their rendering pipelines are complex and inefficient. 
To address these gaps, our dataset provides richly annotated indoor scenes with realistic transparent objects (e.g., containers, cosmetics), complete with background depth maps. It is designed to support scalable, efficient training for future embodied intelligence applications.

\begin{figure*}[htbp]
\centering
\includegraphics[width=0.7\linewidth]{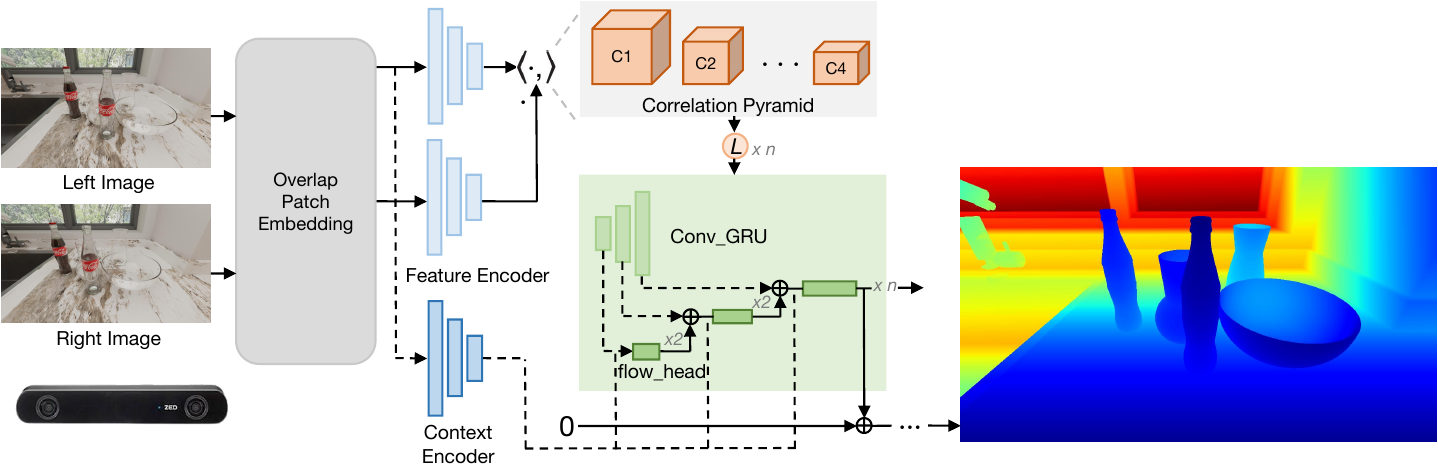}
\caption{
Our stereo depth recovery network for transparent objects. The feature encoder extracts appearance features from both left and right images, while a context encoder processes the left image to provide structural priors for disparity refinement. A correlation pyramid is then constructed by merging left–right features to capture correspondence cues. These features, together with structural priors, are iteratively refined through a GRU-based update loop, which integrates texture similarity and structural consistency. The network finally outputs a refined disparity map that is robust to transparency-induced ambiguities.
}
\label{fig.cleardepthnetwork}
\end{figure*}

\section{Problem Statement and Methods}\label{sec:method}

\subsection{Problem Statement}
Stereo depth estimation for transparent objects is fundamentally challenging because the observed pixel intensity is not solely determined by the surface geometry but also influenced by background refraction and reflection. 
In other words, the imaging process of transparent objects often mixes optical information from the background, making traditional appearance- or texture-based stereo matching inherently ill-posed.

Formally, the intensity $I(x)$ at pixel $x$ can be expressed as:
\begin{equation}
I(x) = \alpha \, T(x) + (1-\alpha) \, B(x),
\end{equation}
where $T(x)$ denotes the transmitted (refracted) signal, $B(x)$ represents the background contribution, and $\alpha \in [0,1]$ is the transparency coefficient. 

To address this limitation, ClearDepth incorporates \textbf{structural features} that are more invariant to transparency effects. Given a disparity field $d(x)$, the stereo matching objective can be formulated as:
\begin{equation}
\min_{d(x)} \; \| I_L(x) - I_R(x-d(x)) \|^2 + \lambda \, \mathcal{R}(d(x), \phi_s(x)),
\end{equation}
where $I_L, I_R$ denote the stereo image pair, and $\mathcal{R}(d(x), \phi_s(x))$ is a structural regularizer that enforces consistency between disparity and structural embeddings $\phi_s(x)$. These embeddings are extracted via a cascaded ViT backbone, whose global self-attention mechanism captures \textbf{long-range shape and contour information}, thereby reducing ambiguities in transparent regions. The cascaded vision transformer backbone is detailed in Sec.~\ref{sec:vision_transformer_backbone}. Furthermore, to compensate for the failure modes of traditional dot-product similarity, we also introduce a \textbf{post-fusion mechanism} that combines texture-based and structure-based disparity estimates:
\begin{equation}
d_f(x) = w_s(x) \, d_s(x) + w_a(x) \, d_a(x), \quad w_s(x) + w_a(x) = 1,
\end{equation}
where $d_a(x)$ is the disparity derived from appearance similarity, $d_s(x)$ is the structure-guided disparity, and $w_s(x), w_a(x)$ are adaptive confidence weights. Intuitively, when texture cues are reliable (opaque regions), $w_a(x)$ dominates, while in transparent or textureless regions, $w_s(x)$ dominates, enforcing structural consistency. The fusion design is introduced in Sec.~\ref{sec:pose_fusion}.

In summary, we propose \textbf{ClearDepth}, employ a ViT backbone for robust structural feature extraction and design a post-fusion module to explicitly compensate for the shortcomings of traditional stereo matching in transparent-object scenarios. Our network is illustrated in Fig.~\ref{fig.cleardepthnetwork}. The dataset generation is presented in Sec.~\ref{sec:synthetic_dataset_generation}.

\subsection{Cascaded Vision Transformer Backbone.}
\label{sec:vision_transformer_backbone}

Our backbone begins with overlap patch embedding for initial tokenization, preserving local features. Tokens pass through four transformer blocks, generating feature maps at $\frac{1}{4}$, $\frac{1}{8}$, $\frac{1}{16}$, and $\frac{1}{32}$ scales. To optimize computational efficiency, the model incorporates efficient self-attention, which significantly reduces the computational burden from $O(N^2)$ to $O(\frac{N^2}{R})$. This reduction is achieved by first reshaping the input sequence from $N \cdot C$ to $\frac{N}{R} \times (C \cdot R)$ by 2d convolutional layer with the stride $8,4,2,1$ for different ViT blocks, as detailed in Equ.~\ref{eq:reshape}, and then adjusting the sequence dimensions back to $C$ channel through linear layers, as described in Equ.~\ref{eq:linear}. K denotes the sequence in the ViT block optimized for lower computational complexity.
\begin{align}
\hat{K}=Reshape(\frac{N}{R},C\cdot R)(K)  \label{eq:reshape} \\
K=Linear(C\cdot R,C)(\hat{K})  \label{eq:linear}
\end{align}
Additionally, the Mix-FFN module in the architecture addresses the challenge of performance degradation due to the interpolation of positional embeddings in the original ViT structure, especially when dealing with varying input image sizes, here by substituting positional embeddings with learnable depth-wise convolutions. The equation is as~\ref{eq:mixffn}. 
\begin{align}
\mathbf{x}_{\text{out}} = \text{MLP}(\text{GELU}(\text{Conv}_{3 \times 3}(\text{MLP}(\mathbf{x}_{\text{in}})))) + \mathbf{x}_{\text{in}}
  \label{eq:mixffn}
\end{align}

Then, we concatenate multi-scale feature maps from different ViT blocks by upsampling them to a unified scale of $\frac{1}{4}$. This combined feature map undergoes further refinement through a precise $1 \cdot 1$ convolution, facilitating optimal dimension adjustment.

\begin{figure*}
    \centering
    \begin{subfigure}[b]{0.7\linewidth}
    \includegraphics[width=\linewidth]{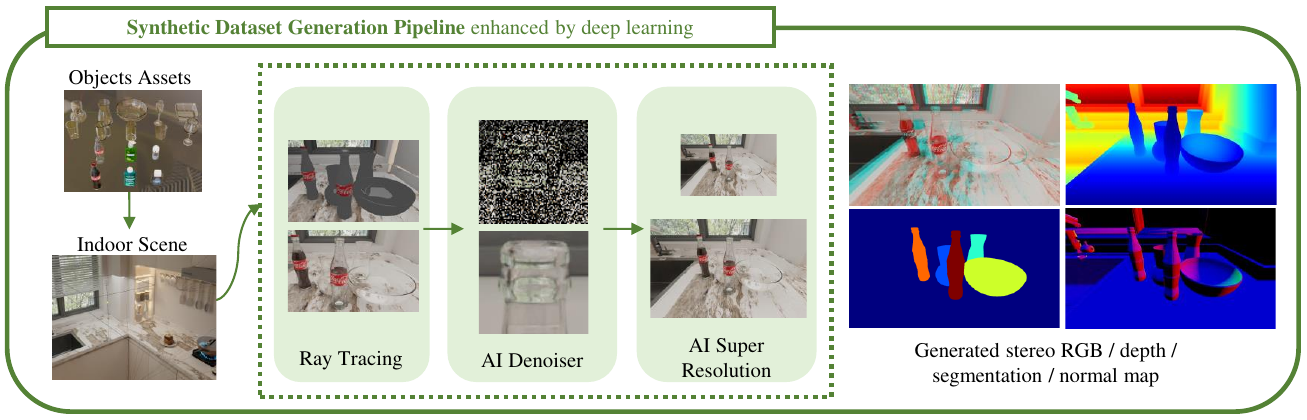}
        \caption{Pipeline of synthetic dataset generation.}
        \label{fig.syndataset_generation_pipeline}

    \end{subfigure}
    \begin{subfigure}[b]{0.13\linewidth}
    \includegraphics[width=\linewidth]{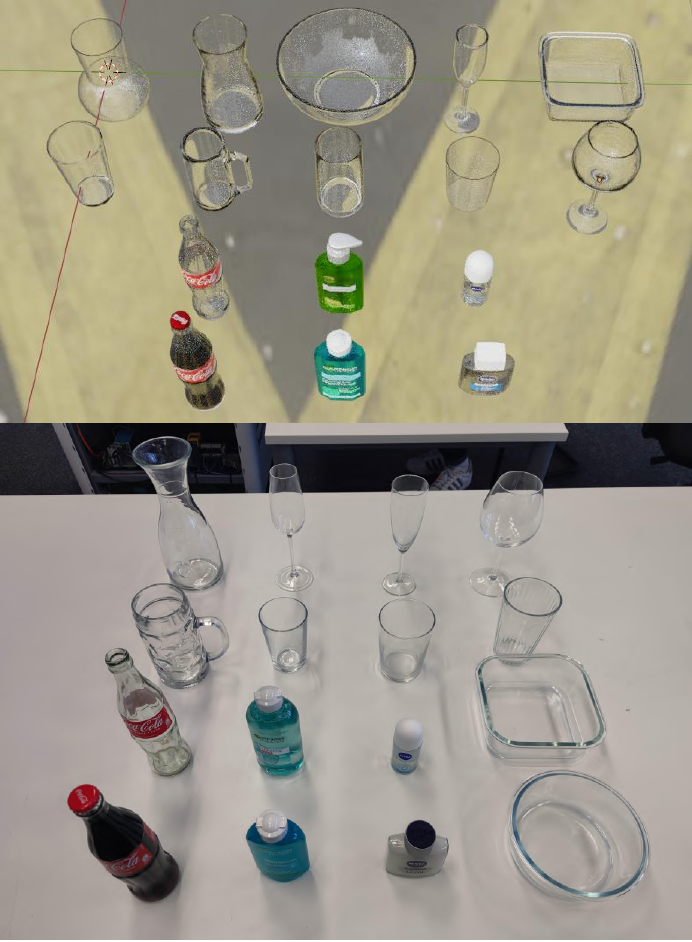}
        \caption{Rendered 3D models and real-world objects}
        \label{fig.cleardepth_syn_and_real_object_database}
    \end{subfigure}
    \\
    \vspace{3pt} 
    \caption{SynClearDepth dataset with diverse objects, various scene configurations.}
\end{figure*}

\subsection{Structural Feature Post-Fusion}
\label{sec:pose_fusion}

We propose a modified GRU-based architecture that refines disparity maps in a coarse-to-fine manner. The Post-Fusion mechanism is specifically designed to address the unique challenges of transparent objects. Our experiments show that, unlike opaque objects, accurate depth estimation of transparent objects relies heavily on fine-grained structural cues. In addition, the refractive nature of transparent surfaces distorts background textures, making dot-product–based feature similarity unreliable for reconstruction. To mitigate this, we incorporate structural information from the image itself into the GRU iterations. This integration ensures that structural cues extracted at multiple resolutions are consistently preserved throughout the iterative refinement process.

The core update equations in our model are defined as follows:
\begin{align}
x_k = & \;[\mathbf{C}_k, \mathbf{d}_k, \mathbf{c}_k, \mathbf{c}_r, \mathbf{c}_h] \\
z_k = & \;\sigma(\text{Conv}([h_{k-1}, x_k], W_z) + c_k), \\
r_k = & \;\sigma(\text{Conv}([h_{k-1}, x_k], W_r) + c_r), \\
\Tilde{h}_k = & \,\tanh(\text{Conv}([r_k \odot h_{k-1}, x_k], W_h) + c_h), \\
h_k = & \;(1-z_k) \odot h_{k-1} + z_k \odot \Tilde{h}_k,
\end{align}

Here, $x_k$ is a concatenation of several feature maps, including the correlation $\mathbf{C}_k$, the current disparity $\mathbf{d}_k$, and structural context feature maps $\mathbf{c}_k$, $\mathbf{c}_r$, and $\mathbf{c}_h$. Specifically, $\mathbf{c}_k$, $\mathbf{c}_r$, and $\mathbf{c}_h$ represent structural features derived from the left image. These features are incorporated as residuals into the GRU loop, allowing for enhanced participation of structural information during the disparity map refinement process. $z, r, h$ represent the state information of the update gate, reset gate, and hidden gate in a GRU.

Then, Our approach decode GRUs at each resolutions to obtain multi-scale disparity updates for coarse to fine gradual optimization:
\begin{alignat}{2}
&\triangle \mathbf{d}_{k,\frac{1}{32}} &&= \text{Decoder}(h_{k,\frac{1}{32}}), \\
&\triangle \mathbf{d}_{k,\frac{1}{16}} &&= \text{Decoder}(h_{k,\frac{1}{16}} + \text{Interp}(\triangle \mathbf{d}_{k,\frac{1}{32}})), \\
&\triangle \mathbf{d}_{k,\frac{1}{8}}  &&= \text{Decoder}(h_{k,\frac{1}{8}} + \text{Interp}(\triangle \mathbf{d}_{k,\frac{1}{16}})),
\end{alignat}
where $\text{Decoder}$ consist of two convolutional layers and $\text{Interp}$ is bilinear interpolation scaled up by a factor of two.
Finally, the updated disparity is calculated as:
\begin{equation}
\begin{aligned}
\mathbf{d}_{k+1} = & \; \mathbf{d}_{k} + \triangle \mathbf{d}_k
\end{aligned}
\end{equation}

In summary, to address the challenges of transparent objects, we selected an appropriate image feature extractor. Additionally, considering the unique difficulties of transparent objects and the need for efficient models in robotics, we designed a structural feature post-fusion architecture. Every detail of our network structure is tailored to the characteristics of transparent object scenarios.

In the comparative experiments section, the visual results demonstrate that our model substantially enhances the stereo imaging of transparent objects.

\subsection{Synthetic Dataset Generation}
\label{sec:synthetic_dataset_generation}

To enhance the efficiency of synthetic dataset generation, we utilized the AI denoiser provided by OptiX~\cite{chaitanya2017interactive} during rendering and adopted open-source pretrained deep learning super-resolution~\cite{ahn2018fast} as rendering output optimization strategies, reducing the average generation time per set (stereo RGB, depth, masks, and object-camera poses) from $12.77$ to $4.40$ seconds. Since these techniques are widely used in computer graphics and do not alter the core data distribution, we omit further analysis. The dataset generation process is illustrated in Fig.~\ref{fig.syndataset_generation_pipeline}.
Our SynClearDepth dataset includes 16 selected objects: 10 common transparent containers and 6 glass-material products (Fig.~\ref{fig.cleardepth_syn_and_real_object_database}). To ensure depth labels for both objects and backgrounds, we combined object models with indoor scenes, including 6 bathrooms, 3 dining rooms, 5 kitchens, and 6 living rooms. This resulted in 14,091 image sets, each containing left and right RGB images, ground truth depth, instance segmentation, and object/camera poses (Fig.~\ref{fig.cleardepth_dataset_image_samples}). We applied domain randomization to object types, quantities, poses, lighting, and camera angles. This dataset is designed to support robotic perception and manipulation in service robot applications, particularly for handling transparent objects in household environments. For more
details, please refer to the supplementary
materials.

\section{Experiments}\label{sec:experiments}

\subsection{Technical Specifications}

Our network is firstly pre-trained on CREStereo dataset~\cite{li2022practical} and Scene Flow dataset~\cite{mayer2016large}, and then fine-tuned on our proposed SynClearDepth dataset for transparent object stereo imaging. Our model is trained on 1 block of NVIDIA RTX A6000 with batch size 8 and the whole training lasts for 300,000 steps. We use AdamW~\cite{loshchilov2019decoupled} as optimizer, the learning rate is set to 0.0002, updated with a warm-up mechanism and used one-cycle learning rate scheduler. The final learning rate when training finished is 0.0001. The input size of the model is resized to $360 \times 720$. Fine-tune for transparent objects takes the same training parameters as pretraining on the opaque dataset.

\subsection{Evaluation Metrics}

\begin{enumerate}
    \item \textbf{AvgErr (Average Error):} Represents the average disparity error across all pixels, indicating the general accuracy of the disparity map.
    \item \textbf{RMS (Root Mean Square Error):} Measures the square root of the average squared disparity error, reflecting the overall deviation from the ground truth.
    \item \textbf{Bad 0.5 (\%), Bad 1.0 (\%), Bad 2.0 (\%), Bad 4.0 (\%):} These metrics indicate the percentage of pixels where the disparity error exceeds 0.5, 1.0, 2.0, and 4.0 pixels, respectively, highlighting the proportion of significant errors in the disparity map.
\end{enumerate}

Together, these metrics provide a comprehensive assessment of stereo matching performance, balancing both overall accuracy and the frequency of large errors.

\begin{table}[htp]
\centering
\caption{Quantitative results on transparent object dataset compared with stereo SOTA methods fine-tuned with SynClearDepth dataset. Visualization results shown in Fig.~\ref{fig.clearDepth-eval-vis}.}
\label{table.evaluation_transparent_dataset}
\resizebox{1.00\linewidth}{!}{
\begin{tabular}{lcccccc}
\toprule
Methods    & AvgErr~$\downarrow$ & RMS~$\downarrow$ & bad 0.5 (\%)~$\downarrow$ & bad 1.0 (\%)~$\downarrow$ & bad 2.0 (\%)~$\downarrow$ & bad 4.0 (\%)~$\downarrow$ \\ 
\hline
IGEV-Stereo\cite{xu2023iterative} & 2.077 & 5.5301 & 58.9743 & 36.1270 & 19.967 & 10.777 \\ 
DLNR\cite{zhao2023high} & 3.097 & 8.4269 & 28.1088 & 21.8442 & 16.481 & 11.9046 \\ 
Selective-IGEV\cite{wang2024selective} & \textbf{1.273} & \textbf{4.3365} & 34.8229 & 17.6288 & \textbf{9.561} & 5.8707 \\ 
RAFT-Stereo\cite{lipson2021raft} & 2.245 & 8.8016 & 29.7356 & 17.4521 & 10.835 & 6.2107 \\ 
\rowcolor[gray]{0.8}
\textbf{ClearDepth (ours)} & 2.138 & 8.7282 & \textbf{24.7329} & \textbf{16.3178} & 9.8459 & \textbf{5.7600} \\ 
\midrule
\end{tabular}
}
\end{table}

\subsection{Qualitative and Quantitative Studies for Stereo Depth Estimation}
\subsubsection{Evaluation on Transparent Object Dataset}

\begin{figure*}[htbp]
\centering
\includegraphics[width=0.9\linewidth]{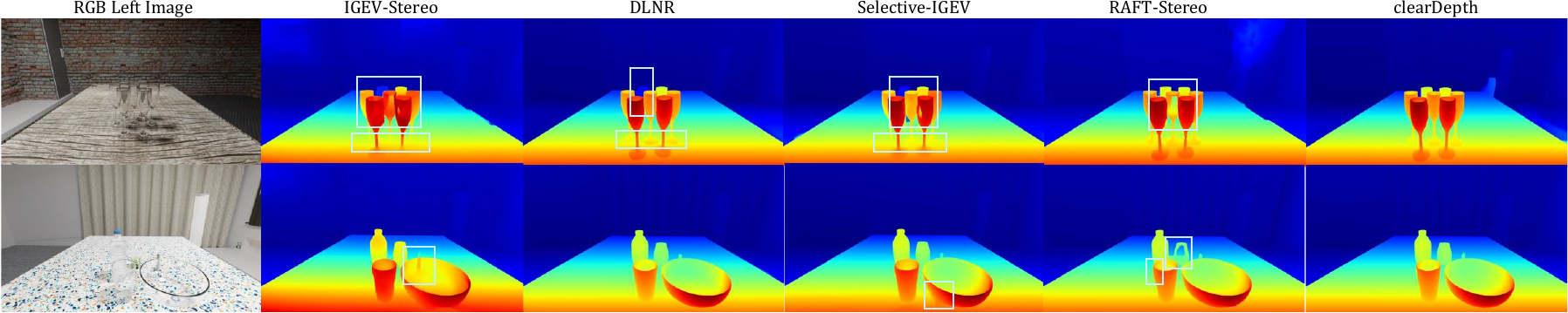}
\caption{
The visualization results of our transparent object stereo depth reconstruction method compare with other SOTA stereo depth estimation methods by fine-tuning on SynClearDepth dataset.}
\label{fig.clearDepth-eval-vis}
\end{figure*}

\begin{figure}
    \centering
        \includegraphics[width=0.8\linewidth]{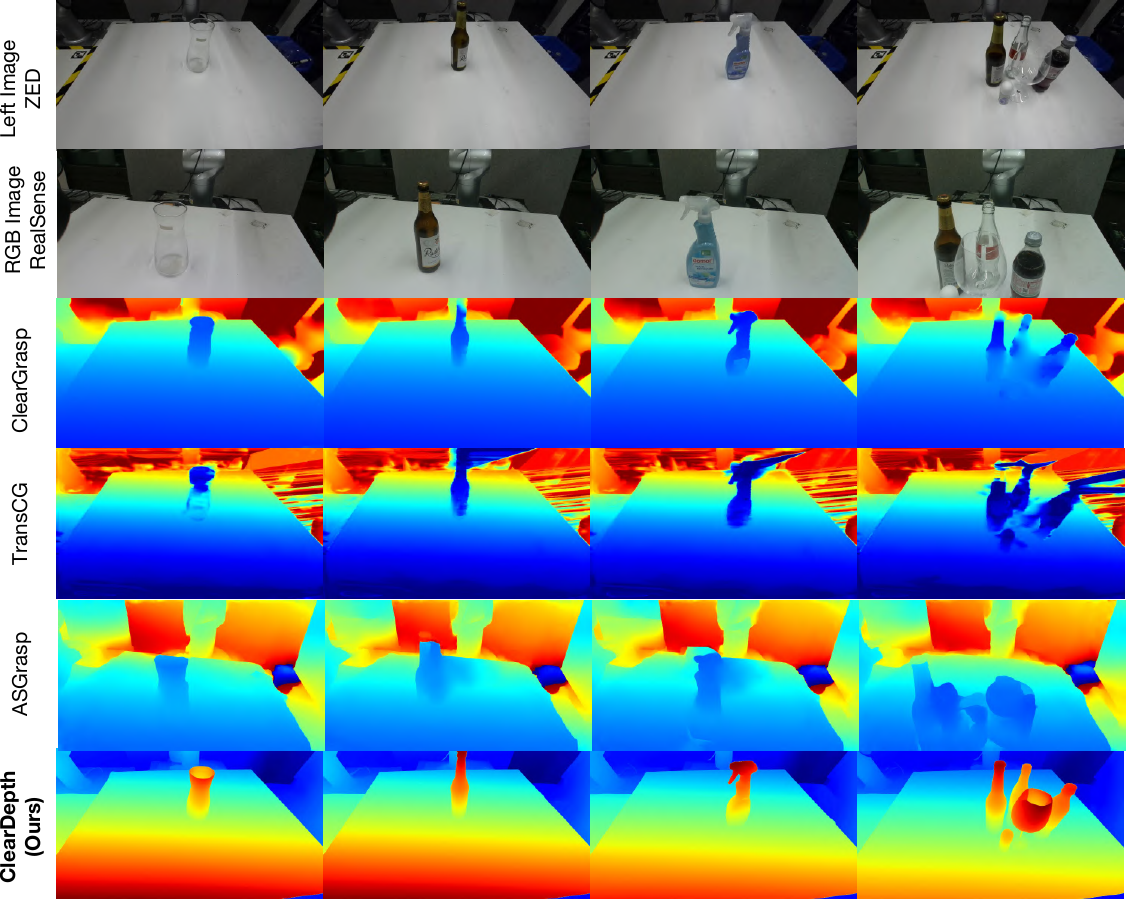}
    \\
    \vspace{3pt} 
        \caption{Qualitative experiments of ClearGrasp~\cite{sajjan2020clear}, TransCG~\cite{fang2022transcg}, ASGrasp~\cite{shi2024asgrasp} and proposed ClearDepth for objects with different materials in single-object and cluttered scene.}
        \label{fig:exp_real_world_different_material}
\end{figure}

\begin{figure}
    \centering
        \includegraphics[width=0.8\linewidth]{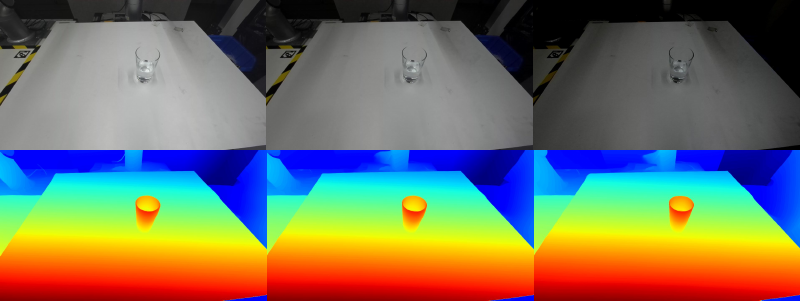}
    \\
    \vspace{3pt} 
        \caption{Qualitative experiments of proposed ClearDepth for scenes with different lighting conditions.}
        \label{fig:exp_real_world_different_lighting_condition}
\end{figure}

To validate our model and dataset for transparent object depth recovery in stereo vision, we fine-tuned our pre-trained model on the SynClearDepth dataset using the same training parameters as pre-training. We also fine-tuned RAFT-Stereo~\cite{lipson2021raft}, IGEV-Stereo\cite{xu2023iterative}, DLNR\cite{zhao2023high}, Selective-IGEV\cite{wang2024selective} from the Middlebury benchmark on SynClearDepth for comparison. This analysis highlights our model’s improvements in stereo-based depth perception for transparent objects. Tab.~\ref{table.evaluation_transparent_dataset} presents quantitative results, while Fig.~\ref{fig.clearDepth-eval-vis} visualizes stereo imaging performance. AvgErr (Average Error) and RMS (Root Mean Square Error) measure numerical error, while Bad 0.5 (\%), Bad 1.0 (\%), Bad 2.0 (\%), Bad 4.0 (\%) reflect relative error. Results show our model achieves strong performance in numerical error and outperforms all others in relative error. 
Our model is more efficient than others, achieving comparable performance without the high computational cost of cross-attention or multi-model ensembles. This is due to innovations in the image encoder, making our approach more suitable for robotics.

\subsubsection{Comparison experiments with SOTA zero-shot stereo matching methods}
To evaluate the effectiveness of our method on transparent object stereo depth estimation, we conduct a comprehensive comparison against several SOTA open-source zero-shot stereo matching approaches~\cite{wang2024selective,zhao2023high,xu2023iterative,wen2025foundationstereo,jiang2025defom} on a dedicated transparent-object validation set. Specifically, we include FoundationStereo~\cite{wen2025foundationstereo}, and DEFOM-Stereo~\cite{jiang2025defom}, all of which claim to generalize to arbitrary unseen scenes without requiring additional training. For fair comparison, we directly adopt their officially released pretrained models and evaluate them under identical conditions involving transparent objects.

Given that our target application is robotic grasping, where the foreground regions (i.e., the object areas) are of primary importance, we compute all quantitative metrics exclusively on these regions to better reflect each model’s performance on the most critical parts of the scene. As shown in Table~\ref{tab:transparent_sota}, our method substantially outperforms all competing methods across all evaluation metrics. It achieves lower average error, reduced root mean square (RMS) error, and the lowest bad-pixel rates under multiple threshold settings. These results demonstrate that our approach not only generalizes effectively to novel transparent-object scenes but also delivers substantial accuracy improvements over existing zero-shot stereo methods.

This also indicates that the lack of transparent-object stereo datasets in the current open-source community negatively impacts the performance of zero-shot stereo methods, further highlighting the value and contribution of our dataset.

\begin{table}[htp]
  \centering
    \caption{Quantitative results on transparent object dataset compared with stereo SOTA zero-shot stereo reconstruction methods. 
  }
  \label{tab:transparent_sota}
  \resizebox{1.0\linewidth}{!}{
    \begin{tabular}{lcccccc}
      \toprule
   Methods    & AvgErr~$\downarrow$ & RMS~$\downarrow$ & bad 0.5 (\%)~$\downarrow$ & bad 1.0 (\%)~$\downarrow$ & bad 2.0 (\%)~$\downarrow$ & bad 4.0 (\%)~$\downarrow$ \\ 
\hline
IGEV-Stereo~\cite{xu2023iterative}        & 26.8047 & 39.0820 & 0.968369 & 0.939858 & 0.890477 & 0.783732 \\
DLNR~\cite{zhao2023high}               & 26.5240 & 38.0410 & 0.962301 & 0.927577 & 0.865248 & 0.758555 \\
Selective-IGEV~\cite{wang2024selective}     & 23.8168 & 36.2975 & 0.959417 & 0.921393 & 0.854298 & 0.731494 \\
RAFT-Stereo~\cite{lipson2021raft}     & 29.2919 & 39.2978 & 0.971589 & 0.946811 & 0.901663 & 0.821213 \\
DEFOM-Stereo~\cite{jiang2025defom}       & 16.1635 & 25.8188 & 0.890889 & 0.792165 & 0.680660 & 0.550519 \\
FoundationStereo~\cite{wen2025foundationstereo}   & 8.8985  & 16.0874 & 0.891110 & 0.799528 & 0.668938 & 0.498191 \\
\rowcolor[gray]{0.8}
\textbf{ClearDepth (ours)} & \textbf{3.1084}  & \textbf{6.9570}  & \textbf{0.806759} & \textbf{0.631477} & \textbf{0.375651} & \textbf{0.153726} \\
\midrule
    \end{tabular}
    }
  \vspace{3pt}    
\end{table}

\begin{table}[h!]
  \centering
  \caption{Ablation study for the feature post-fusion module in clearDepth with 100,000 steps on SynClearDepth dataset.}
  \label{table.ablation_transparent_dataset}
  \resizebox{1.0\linewidth}{!}{
    \setlength{\tabcolsep}{3.pt}
    \begin{tabular}{lcccccc}
      \toprule
        Methods & AvgErr~$\downarrow$ & RMS~$\downarrow$ & bad 0.5 (\%)~$\downarrow$ & bad 1.0 (\%)~$\downarrow$ & bad 2.0 (\%)~$\downarrow$ & bad 4.0 (\%)~$\downarrow$ \\
      \hline
      w/o Fusion & 6.90 & 15.48 & 43.34 & 29.63 & 21.52 & 16.62\\
      \rowcolor[gray]{0.8}
      Feature Fusion  & \bf{2.64} & \bf{8.59} & \bf{27.23} & \bf{16.87} & \bf{11.28} & \bf{7.72}\\
      \midrule
    \end{tabular}}
  \vspace{3pt}    
  
\end{table}

\subsubsection{Ablation Study of Feature Post-Fusion Module}

To evaluate the impact of our feature post-fusion module, we conducted ablation studies on the SynClearDepth dataset. We compared networks with and without the module, as shown in Tab.~\ref{table.ablation_transparent_dataset}. Results indicate a substantial performance boost, especially in handling complex transparency and light refraction, highlighting its effectiveness in enhancing depth estimation and object recognition. Each study was trained for 100,000 steps.

\subsubsection{Qualitative experiments on real-world scenes with different materials, lighting conditions}
We perform qualitative analysis on real-world images with different materials and lighting conditions, as shown in Fig.~\ref{fig:exp_real_world_different_material} and Fig.~\ref{fig:exp_real_world_different_lighting_condition}. 
We compare depth perception performance for objects with different materials using our method with SOTA methods, including ClearGrasp~\cite{sajjan2020clear}, TransCG~\cite{fang2022transcg}, ASGrasp~\cite{shi2024asgrasp}. 
For more results, please check out our supplementary materials and videos. 
Results in supplementary video show that leveraging a physically realistic renderer enables strong generalization in real world, with performance consistent across domain-shifted test sets. 
In this work, we adopt a stereo-based approach instead of the Realsense-based imaging methods~\cite{shi2024asgrasp}.  
The open-source datasets using Realsense cameras are limited in both diversity and scale compared to stereo datasets, restricting future extensions. 
Additionally, Realsense IR projection penetrates transparent objects, leading to visual information loss~\cite{shi2024asgrasp}, which we avoid to ensure robustness.
\subsection{Trade-off between Speed and Accuracy}
\subsubsection{Comparison experiment of inference speed and FLOPs}
We compare the speed and average error of our method and SOTA methods, as shown in Fig.~\ref{fig.cleardepth_overview}. For detailed data, please refer to the supplementary material.

\subsubsection{TensorRT implementation}
Additionally, our TensorRT implementation enables real-time inference at 50 FPS on consumer GPUs, whereas other models, due to their complex designs, are impractical for deployment.

\begin{table}[h!]
  \centering
    \caption{Real-world robotic grasping comparison experiments with SOTA methods for transparent objects.}
\label{table.grasping_success_rate_sota_comparison}
  \resizebox{1.0\linewidth}{!}{
    \setlength{\tabcolsep}{3.pt}
    \begin{tabular}{lcccc}
      \toprule
        Grasp SR & single (L1) & cluttered (L1) & single (L2) & cluttered (L2) \\
      \hline
  Baseline~\cite{stereolabs_zed2} & 78\% & 63\% & 62\% & 58\% \\
TransCG~\cite{fang2022transcg} & 80\% & 70\% & 78\% & 67\% \\
\rowcolor[gray]{0.8}
\textbf{ClearDepth (ours)} & 98\% & 92\% & 98\% & 90\% \\
      \midrule
    \end{tabular}
    }
  \vspace{3pt}    

\end{table}

\begin{figure}[htbp]
\centering
\includegraphics[width=0.7\linewidth]{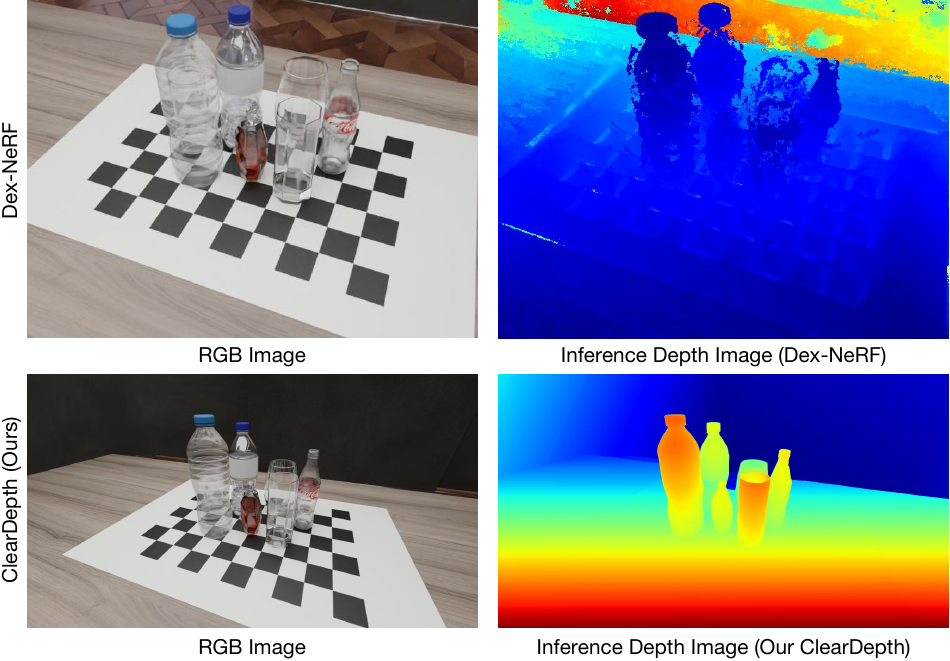}
\caption{Comparison experiment with NeRF-based methods~\cite{ichnowski2021dex}.
}
\label{fig.comparison_exp_with_nerf}
\end{figure}

\begin{figure}[htbp]
\centering
\includegraphics[width=1.0\linewidth]{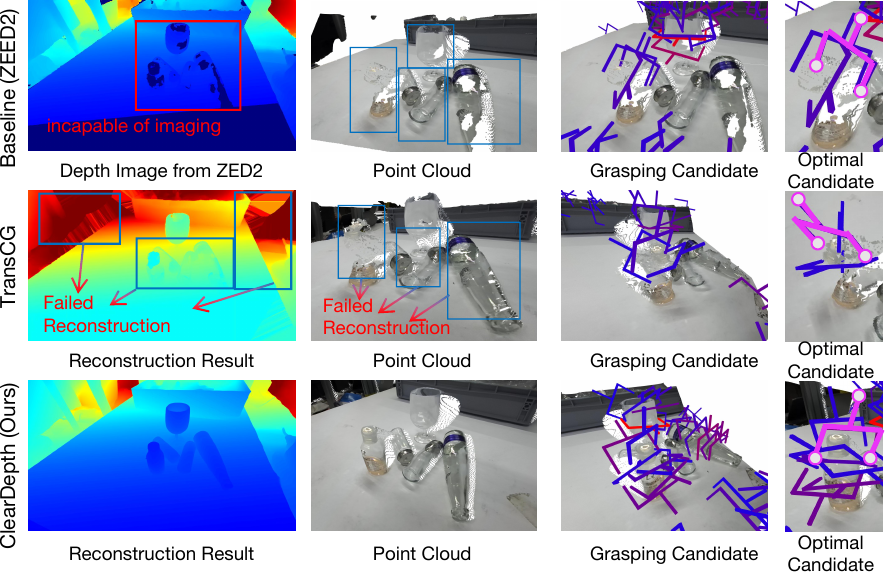}
\caption{Real-world qualitative comparisons of transparent object grasping using depth reconstruction of ZED2~\cite{stereolabs_zed2}, TransCG~\cite{fang2022transcg}, our ClearDepth. The grasping candidates are estimated using GraspNet-Baseline~\cite{fang2023robust}. Depth images, point clouds, and grasping results are presented.
}
\label{fig.comparison_exp_with_grasping_sota}
\end{figure}

\subsection{Comparison experiment with NeRF-based method}
We execute comparison experiment with NeRF-based method~\cite{ichnowski2021dex}. The reconstructed depth images are shown in Fig.~\ref{fig.comparison_exp_with_nerf}. 
Our method achieves better reconstruction quality compared to NeRF-based method~\cite{ichnowski2021dex}, which requires additional data acquisition and suffers from low efficiency. In the context of robotic manipulation tasks, training a separate model for each scene introduces considerable overhead.

\subsection{Additional experiments}
We execute addifional comparison experiments with~\cite{lipson2021raft,guo2023openstereo,li2022practical} in Middleburry dataset~\cite{middlebury_stereo} and KITTI dataset, as detailed in Supplementary Materials.

\subsection{Comparison Experiments of Transparent Object Grasping}
\begin{figure}[htbp]
\centering
\includegraphics[width=0.8\linewidth]{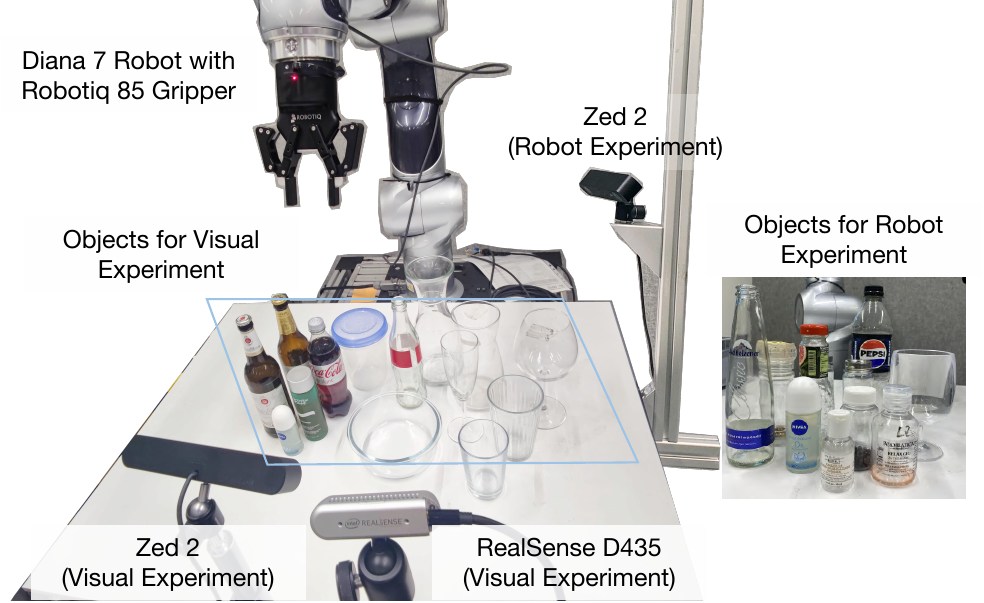}
\caption{Experiment setup for grasping comparison experiment.
}
\label{fig.comparison_exp_setup_two_jaw}
\end{figure}

To evaluate the performance of our transparent object grasping pipeline compared to state-of-the-art (SOTA) methods~\cite{fang2022transcg}, we conducted real-world experiments involving two-finger grasps on transparent objects, as shown in Fig.~\ref{fig.comparison_exp_setup_two_jaw}. The depth data based on the stereo reconstruction method~\cite{stereolabs_zed2} from the ZED camera is used for grasp generation as a baseline method. 
The evaluation scenarios include both single-object grasping and grasping in cluttered environments. Specifically, Level-1 (L1) scenes contain a mix of transparent and opaque objects, while Level-2 (L2) scenes consist exclusively of fully transparent objects. 
For each experimental setting, we performed 150 grasping trials. The grasp success rate is computed as the number of successful grasps divided by the total number of attempts. 
The grasp success rates for all methods are summarized in the Tab.~\ref{table.grasping_success_rate_sota_comparison}, and the corresponding reconstruction results and grasp predictions are illustrated in the Fig.~\ref{fig.comparison_exp_with_grasping_sota}. 
Our method consistently achieves the highest performance across all levels of scene and object complexity. Specifically, it demonstrates superior grasp success rates in both single-object and multi-object scenarios.

\subsubsection{Analysis of Robotic Grasping Experiments}
\begin{figure}[htbp]
\centering
\includegraphics[width=0.7\linewidth]{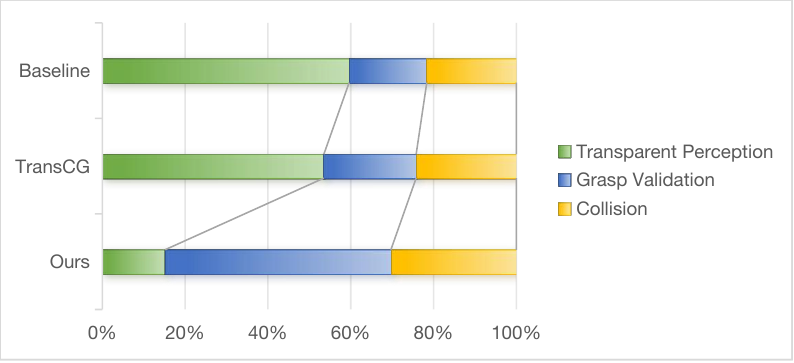}
\caption{Error distribution of baseline method, TransCG~\cite{fang2022transcg} and our ClearDepth. 
We compare the proportions of total failures represented by different failure types.}
\label{fig.error_analysis}
\end{figure}

To evaluate the effectiveness of our method, we conducted an in-depth analysis of the causes of grasp failures. 
The primary cause of failure lies in inaccurate depth reconstruction, which directly leads to unsuccessful grasp attempts. Additionally, grasp prediction errors may also result in collisions or object drops during execution. 
Specifically, the limitations of depth reconstruction manifest in two ways: (1) the inability to perceive transparent regions, leading to collisions between the gripper and the object during execution; and (2) the prediction of noisy points within transparent regions, causing grasp candidates to be located in unreliable areas, ultimately resulting in failure.
The distribution of failure causes across different methods is shown in Fig.~\ref{fig.error_analysis}. It is evident that our method substantially reduces the proportion of failures caused by inaccurate depth reconstruction to increase the grasping success rate.

\subsection{Multi-Fingered Robotic Grasping Experiment}
We also employ our pipeline for transparent object grasping in a robot platform with a robotic arm and multi-fingered robotic hand, as shown in Fig.~\ref{fig.cleardepth_overview}. 
Following grasping pipeline from ContactDexNet~\cite{zhang2024multi}, multi-fingered robotic grasping experiment 
achieves an 86.2\% average success rate.

\section{Conclusion and Future Work}
In this work, we present a complete visual perception framework for transparent object manipulation in service robotics scenarios, spanning synthetic data generation, stereo depth estimation, and real-world robotic validation. We propose an efficient real-time stereo depth recovery network that combines a cascaded vision transformer backbone with a structural feature post-fusion module, enabling fine-grained structural perception and accurate depth recovery of transparent objects without relying on mask priors. To address the data scarcity challenge in transparent object perception, we construct SynClearDepth, a high-quality simulation dataset containing diverse household environments and realistic object placements. It provides accurate RGB, depth maps, instance masks, and pose annotations, substantially enhancing model generalization in real-world scenarios. 
We validate our model through extensive comparisons on public and proprietary datasets, along with ablation studies. Experimental results demonstrate that our approach outperforms existing methods on both public and proprietary benchmarks, particularly in structure-aware and boundary-level depth estimation. 
Results demonstrate its robustness, accuracy, and efficiency, supporting transparent object manipulation in robotics. Furthermore, real-world robotic grasping experiments show that our method can be seamlessly integrated into grasping pipelines without requiring multi-view capture or additional pre-processing, and achieves stable and precise manipulation of transparent objects. These results highlight the practicality and applicability of stereo-based transparent object depth estimation in real-world robotic tasks.

{
    \small
    \bibliographystyle{IEEEtran}
    \bibliography{root}
}
\newpage

\appendix
This supplemental material mainly contains:
\begin{itemize}
    \item Implementation Details of SynClearDepth Dataset, as described in Sec.~\ref{imp_details_syncleardepth}.
    \item Experimental results on Middlebury Dataset, as detailed in Sec.~\ref{results_Middlebury}. 
    \item Experimental results on KITTI dataset and implementation details, as detailed in Sec.~\ref{result_kitti} and Sec.~\ref{implementation_kitti}.
    \item Detailed result of speed-accuracy test using our method and SOTA methods, as shown in Sec.~\ref{speed-accuracy-test}.
\end{itemize}
\subsection{Implementation Details of SynClearDepth Dataset}
\label{imp_details_syncleardepth}
\begin{figure}[htbp]
\centering
\includegraphics[width=0.9\linewidth]{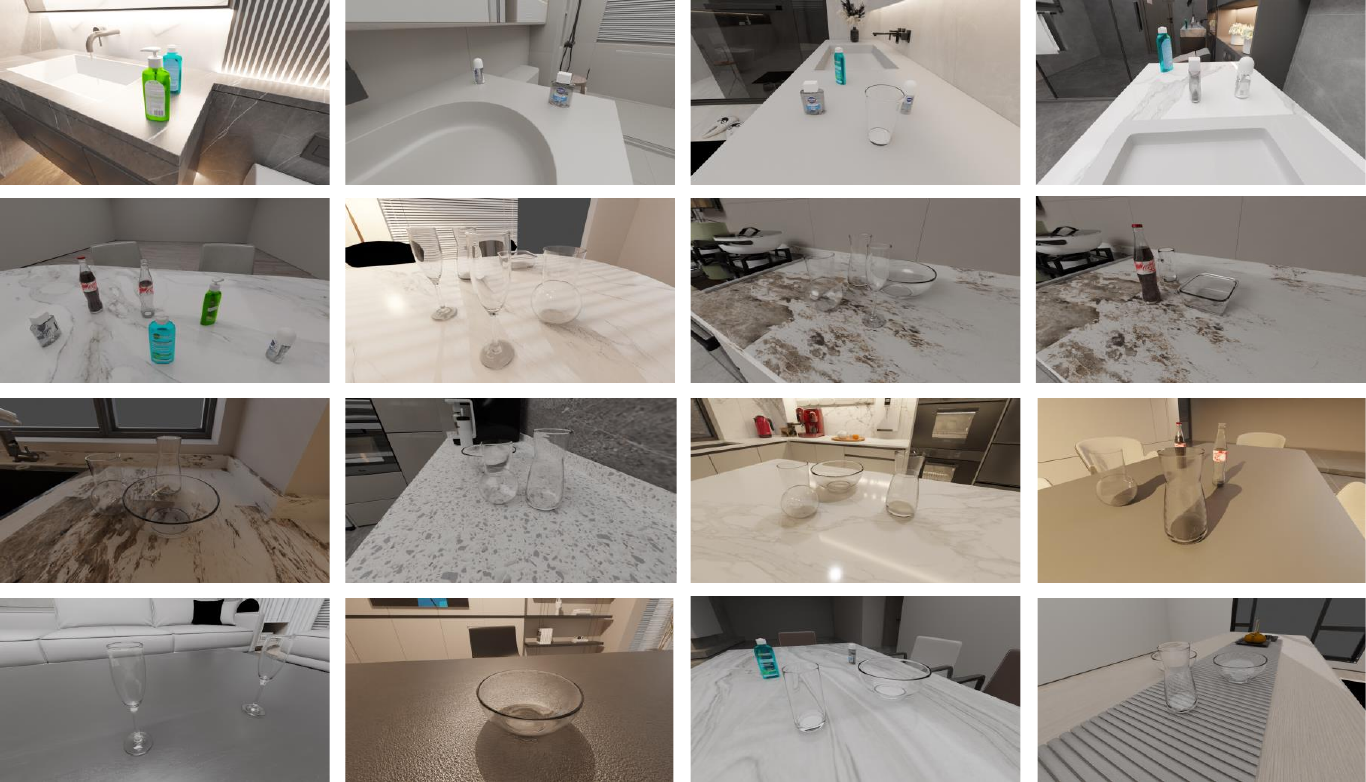}
\caption{
Sample images from SynClearDepth dataset, depicting transparent objects randomly placed in indoor scenes (bathroom, dining room, kitchen, living room) under various lighting conditions. The objects, including cosmetic packaging and glass containers, are randomly dropped onto tables using 3D bounding boxes as collision bodies.
}
\label{fig.cleardepth_dataset_image_samples}
\end{figure}
\subsection{Additional Experiments}
\subsubsection{Quantitative Analysis on Middlebury Dataset}
\label{results_Middlebury}
The Middlebury 2014 dataset comprises 23 pairs of images designate for training and validation purposes. We refine our model over these 23 pairs, conducting fine-tuning across 4,000 iterations with an image resolution of $384 \times 1024$. Benchmark against standard baseline approaches RAFT-Stereo and CREStereo using various stereo evaluation metrics further underscores the efficacy of our approach, as outlined in Tab.~\ref{table.evaluation_middleburry}. More comparison results with other methods can be found at~\cite{middlebury_stereo}.

\subsubsection{Quantitative Analysis on KITTI Dataset}
\label{result_kitti}

\begin{figure}[h]
    \centering
    \begin{subfigure}[b]{0.48\linewidth}
        \includegraphics[width=\linewidth]{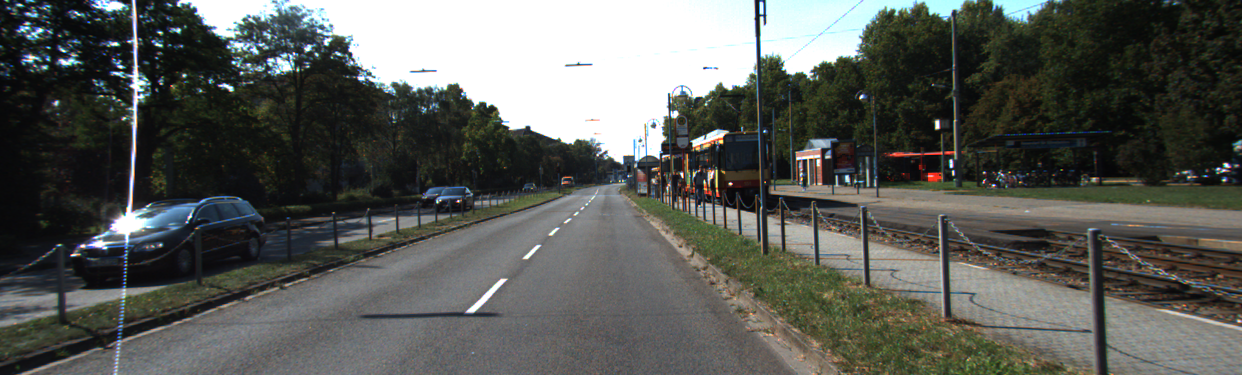}
        \caption{Left image}
    \end{subfigure}
    \begin{subfigure}[b]{0.48\linewidth}
        \includegraphics[width=\linewidth]{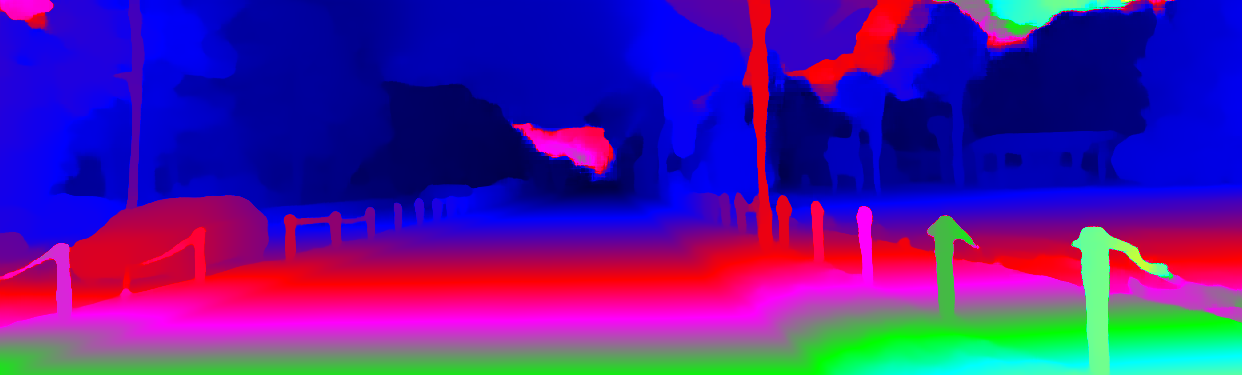}
        \caption{RAFT-Stereo\cite{lipson2021raft}}
    \end{subfigure}
    \begin{subfigure}[b]{0.48\linewidth}
        \includegraphics[width=\linewidth]{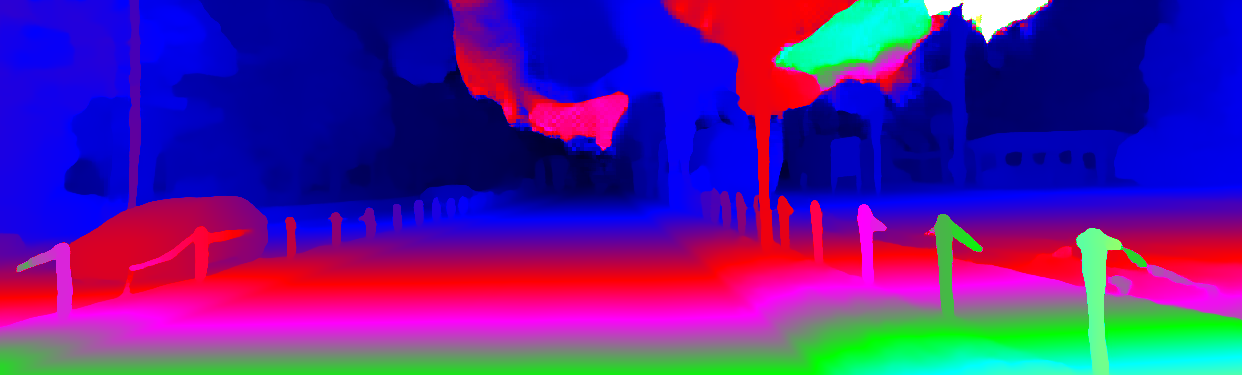}
        \caption{StereoBase\cite{guo2023openstereo}}
    \end{subfigure}
    \begin{subfigure}[b]{0.48\linewidth}
        \includegraphics[width=\linewidth]{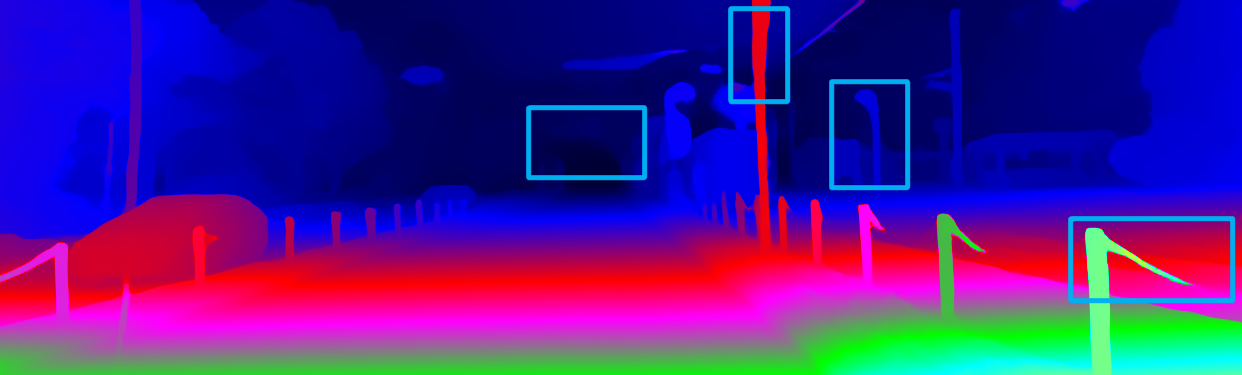}
        \caption{Ours}
    \end{subfigure}
    \\
    \vspace{3pt} 
    \caption{Visual comparisons on KITTI 2015 with SOTA StereoBase and baseline RAFT-Stereo. Our method is more robust to overall scene details.}
    \label{fig:compare_kitti}
\end{figure}

We fine-tune our pre-trained model using the KITTI 2015 training set for comparison experiments with methods~\cite{lipson2021raft,guo2023openstereo}. Given that the labels and metrics of the KITTI dataset cannot fully reflect imaging quality~\cite{depth_anything_v2}, we conducted a qualitative analysis on the KITTI dataset. 
Fig.~\ref{fig:compare_kitti} demonstrates a competitive comparison focused on detail recovery, our method shows exceptional proficiency in reconstructing depth details of foreground objects, significantly outstripping alternative approaches by a substantial margin.

\subsubsection{Implementation Details of Experiments on KITTI Dataset}
\label{implementation_kitti}
We fine-tune our pre-trained model using the KITTI 2015 training set across 5,000 steps, employing image crops sized at $320 \times 1000$. The learning rate is established at 0.00001, with the batch size held at 3. In terms of GRU updates, we perform 22 iterations during training, adjusting to 32 iterations for testing. 

\begin{table}[h]
  \centering
    \caption{Quantitative results on Middleburry Stereo Evaluation Benchmark~\cite{middlebury_stereo}. All metrics have been calculated using undisclosed weighting factors. The outcomes unequivocally demonstrate that our technique substantially outperforms the baseline method.}
  \label{table.evaluation_middleburry}
  \resizebox{1.0\linewidth}{!}{
    \setlength{\tabcolsep}{3.pt}
    \begin{tabular}{lcccccc}
      \toprule
        Methods & AvgErr~$\downarrow$ & RMS~$\downarrow$ & bad 0.5 (\%)~$\downarrow$ & bad 1.0 (\%)~$\downarrow$ & bad 2.0 (\%)~$\downarrow$ & bad 4.0 (\%)~$\downarrow$ \\
      \hline
      RAFT-Stereo\cite{lipson2021raft} & 1.27 & 8.41 & 27.7 & 9.37 & 4.14 & 2.75\\
      CREStereo\cite{li2022practical} & \bf{1.15} & \bf{7.70} & 28.0 & 8.25 & 3.71 & 2.04\\
      \rowcolor[gray]{0.8}
      \textbf{ClearDepth (ours)} & 1.33 & 8.68 & \bf{25.30} & \bf{7.39} & \bf{3.48} & \bf{2.00}\\
      \midrule
    \end{tabular}
    }
  \vspace{3pt}    

\end{table}

\subsection{Detailed Results of Speed-Accuracy Test}
\label{speed-accuracy-test}
The results of speed-accuracy experiment are shown in Tab.~\ref{tab:latency_comparison}.
\begin{table}[htbp]
\centering
\caption{Comparison of Inference Efficiency and Resource Consumption Across Methods}
\label{tab:latency_comparison}
\resizebox{1.0\linewidth}{!}{
    \setlength{\tabcolsep}{3.pt}
\begin{tabular}{lccccc}
\hline
\textbf{Method} & \textbf{Latency (ms)} & \textbf{FPS} & \textbf{FLOPs (G)} & \textbf{Params (M)} & \textbf{GPU Mem (GB)} \\
\hline
CREStereo           & 618.40  & 1.62 & 7056.40 & 5.43   & 1.578 \\
IGEV-Stereo         & 186.66  & 5.36 & 4679.87 & 12.60  & 1.048 \\
DLNR                & 215.31  & 4.64 & 5950.69 & 57.38  & 1.559 \\
Selective-IGEV      & 269.37  & 3.71 & 6220.08 & 13.14  & 1.208 \\
RAFT-Stereo         & 685.78  & 1.46 & 5195.32 & 11.12  & 1.513 \\
DEFORM-Stereo       & 1967.92 & 0.51 & 9607.37 & 382.62 & 6.429 \\
FoundationStereo    & 2050.18 & 0.49 & 23255.41 & 374.52 & 7.089 \\
\rowcolor[gray]{0.8}
\textbf{ClearDepth (Ours)} & 924.88  & 1.08 & 5329.35 & 99.45 & 2.193 \\
\hline
\end{tabular}
}

\end{table}

\end{document}